\documentclass{article}
\usepackage{graphicx} 
\usepackage{caption}
\usepackage{subcaption}
\usepackage{geometry}
\usepackage{blindtext}
\usepackage{amsmath}
\usepackage{amsfonts}
\usepackage{amssymb}
\usepackage{tokcycle}
\usepackage{xcolor}
\usepackage{wrapfig}
\usepackage{pifont}
\usepackage{xifthen}
\usepackage{multirow}
\usepackage{hyperref}
\usepackage{authblk}
\usepackage{algorithm2e}
\usepackage[numbers]{natbib}
\usepackage{tikz}
\usepackage{pgfplots}
\usepackage{scalerel}
\usetikzlibrary{positioning, decorations.pathreplacing,calligraphy, calc, shapes.geometric, intersections, arrows}
\usepgfplotslibrary{fillbetween}
\pgfplotsset{compat=1.18}

\SetKwComment{Comment}{/* }{ */}

\RestyleAlgo{ruled}

\newcommand{\cmark}{\ding{51}}%
\newcommand{\xmark}{\ding{55}}%

\newcommand{\noise}{\sigma}

\title{Deep Recurrence for Dynamical Segmentation Models}
\author[1]{David Calhas}
\author[1]{Arlindo L. Oliveira}
\affil[1]{INESC-ID \\Instituto Superior Tecnico}
\date{}

\begin{document}

\maketitle

\begin{abstract}
    While biological vision systems rely heavily on feedback connections to iteratively refine perception, most artificial neural networks remain purely feedforward, processing input in a single static pass. In this work, we propose a predictive coding inspired feedback mechanism that introduces a recurrent loop from output to input, allowing the model to refine its internal state over time. We implement this mechanism within a standard U-Net architecture and introduce two biologically motivated operations, softmax projection and exponential decay, to ensure stability of the feedback loop. Through controlled experiments on a synthetic segmentation task, we show that the feedback model significantly outperforms its feedforward counterpart in noisy conditions and generalizes more effectively with limited supervision. Notably, feedback achieves above random performance with just two training examples, while the feedforward model requires at least four. Our findings demonstrate that feedback enhances robustness and data efficiency, and offer a path toward more adaptive and biologically inspired neural architectures. Code is available at: \href{https://github.com/DCalhas/feedback_segmentation}{github.com/DCalhas/feedback\_segmentation}.
\end{abstract}

\section{Introduction}

Despite their biological inspiration, most artificial neural networks are static systems: once a prediction is made, these models cannot revise it based on further analysis, a clear contrast with how the brain processes information. The human brain is inherently dynamic: it constantly updates its internal model of the world to minimize discrepancies between predictions and sensory input. This concept is formalized in the \textit{predictive coding theory} \cite{rao1999predictive}, which proposes that the brain functions as a hierarchical inference system. In this framework, higher-level cortical areas generate predictions about sensory input, while lower-level areas compute the error between the prediction and the actual input. This error signal is then used to update the internal state of the system. Inspired by this theory, we propose a feedback mechanism for artificial neural networks that is closely aligned with the predictive coding framework. Our model evolves its internal state by incorporating error signals derived from its own predictions. By feeding the model's output back into itself, we create a dynamical system capable of refining its internal representations over time. This iterative process contrasts with traditional feedforward models, which are limited to a single pass of computation. This iterative refinement process, starting from a blank or uncertain state and evolving towards a structured internal representation, mirrors the way perceptual inference is believed to occur in the brain \cite{rao1999predictive, alamia2021role}. In predictive coding, each step involves reducing the discrepancy between top-down predictions and bottom-up sensory evidence. Crucially, the trajectory toward the correct interpretation is not fixed or known in advance. Instead, the system adapts dynamically, updating its internal model to converge toward the goal state \cite{bogacz2017tutorial}.

We validate our approach on a semantic segmentation task using a modified U-Net \cite{ronneberger2015u}, showing that the incorporation of predictive coding-inspired feedback leads to improved robustness and generalization, especially in noisy or low-data regimes. Our contributions are:

\begin{itemize}
    \item we apply feedback in a widely used neural architecture (see Section \ref{section:methods}) and show its superiority against its original/feedforward version (see Section \ref{section:results});
    \item we propose two operations (see Section \ref{section:methods}) that ensure stability in the feedback system (see Figure \ref{fig:pca_various_feedback}), without impacting the performance of the model (see Table \ref{tab:ablations});
    \item we show that with feedback the system is able to process noisy images better than the purely feedforward system and we are the first to show that it has a better generalization ability when learning from limited samples (see Section \ref{section:discussion} and Figure \ref{fig:feedforward_vs_feedback}).
\end{itemize}

\section{Related work}\label{section:related}

\subsection{Training recurrent neural networks}

Recurrent neural networks \cite{jordan1997serial} are designed to model temporal dependencies by evolving their internal states over time. However, training these models introduces challenges, particularly when applying backpropagation through time (BPTT), which requires storing all intermediate states. This results in high memory consumption. Recurrent backpropagation, originally proposed by \citet{pineda1987generalization} and \citet{almeida1990learning}, mitigates this by allowing the network to evolve until it reaches a steady state, and then computing the gradient at equilibrium. This fixed point approach can significantly reduce memory costs. Later, \citet{liao2018reviving} proposed efficient variants of this algorithm. However, stability remains an issue. For example, \citet{linsley2020stable} introduced a regularization strategy to retain information necessary for training, noting that some gradients are only present in early iterations. Due to the nonlinearity of deep architectures, especially those with feedback, we found recurrent backpropagation to be unstable and difficult to apply to an entire network. Instead, we adopted a strategy that encourages convergence to a stable state quickly and applied standard BPTT over the trajectory, inspired by recurrent backpropagation (force the system to be stable) but using the whole trajectory.

\subsection{Recurrency is biologically plausible}

Biological systems are inherently recurrent \cite{kar2019evidence}. The human brain processes sensory input via loops that connect higher and lower level cortical areas, supporting both feedforward and feedback information flow. This mechanism has inspired models like the Hopfield network \cite{hopfield1982neural}, originally designed for associative memory and later extended with increased capacity \cite{krotov2016dense, hoover2024energy}. More recently, \citet{spoerer2017recurrent} showed that recurrent convolutional models are better at object recognition than feedforward ones, particularly under challenging conditions. Likewise, \citet{kietzmann2019recurrence} demonstrated that recurrence is necessary to capture the dynamic representational signatures observed in the human brain. A prominent line of work aims to connect these computational models to predictive coding theory, the idea that the brain minimizes prediction errors by feeding top down predictions to lower level areas \cite{rao1999predictive}. \citet{alamia2021role} showed that recurrence plays a central role in predictive visual processing, aligning with the predictive coding framework. Additionally, \citet{goetschalckx2024computing} studied how recurrent models can approximate human like uncertainty and reaction time. These works highlight the biological plausibility of recurrence, but they often fall short of explicitly modeling the error correcting loop central to predictive coding.

\subsection{Previous recurrent models and key differences}

While many models incorporate recurrence, few implement the core mechanisms of predictive coding, particularly the idea of dynamically updating internal states through top down prediction and error correction across multiple time steps. Several works adopt recurrence by unrolling the layers of a feedforward network over time, propagating the signal repeatedly through the same sequence of layers. For instance, CORnet \cite{kubilius2018cornet} introduces a recurrent architecture that mimics visual cortex processing stages, and \citet{alamia2021role} study visual predictive processing using a similar unrolled structure. While these models are biologically motivated, their recurrence receives states from adjacent layers: each cortical layer (or its neural network analogue) processes its own input over time and the state of the adjacent following layer, but the architecture lacks direct feedback connections between higher and lower levels. By contrast, our model explicitly connects high level outputs back to earlier layers by concatenating the evolving internal state with the original input. This creates a feedback loop that spans the full abstraction hierarchy, allowing prediction error to influence the representation at every level, a hallmark of predictive coding theory \cite{rao1999predictive}. Additionally, some prior works use recurrence only at the final layers of a network \cite{spoerer2017recurrent, goetschalckx2024computing}, or by repeating layers with shared weights, forming \textit{stateful models} without hierarchical feedback. Our work differs by:
\begin{itemize}
    \item defining a feedback system in which the evolving output is recursively fed back and used to update an internal state;
    \item applying this mechanism across an entire segmentation network (U-Net), not just in final layers;
    \item incorporating operations such as exponential decay and softmax normalization to ensure convergence, stability, and robust representations.
\end{itemize}
Recent advances such as Deep Equilibrium Models (DEQ) \cite{bai2019deep} and Fixed Point Networks (FPN) \cite{heaton2021feasibility} also define implicit recurrent systems, converging to a stable fixed point through root finding or optimization. These models are similar in spirit to our approach as they seek equilibrium solutions rather than standard feedforward activations. However, unlike DEQ or FPN, which either optimize only at equilibrium or apply truncated backpropagation through time, we propose optimizing the entire trajectory. This continuous trajectory based learning is motivated by the hypothesis that the full path contains rich information necessary for robust inference and learning, especially in dynamic or noisy settings. Crucially, our model makes this feasible by ensuring fast convergence through the use of exponential decay, enabling full trajectory training without prohibitive computational cost. This full trajectory optimization avoids the loss of information typically incurred by truncated learning or equilibrium fixed point schemes and culminates in a solution to the challenge identified by \citet{linsley2020stable}.

\section{Methods}
\label{section:methods}

\noindent\textbf{Predictive coding and circuit design.} Predictive coding theory posits that the brain constantly updates its internal representation of the world by minimizing the discrepancy between top down predictions and bottom up sensory input \cite{rao1999predictive}. Inspired by this mechanism, we propose a feedback based model that iteratively refines its internal state to align with the input. We introduce a state vector \( \mathbf{h}(t) \in \mathbb{R}^{d} \) that evolves over time. At each timestep \( t \), the state is updated to reduce the prediction error. The state vector is divided into two disjoint subsets:
\begin{itemize}
    \item \( \mathbf{u}(t) \in \mathbb{R}^{l} \): segmentation neurons, used to produce the predicted mask;
    \item \( \mathbf{v}(t) \in \mathbb{R}^{k} \): feedback neurons, used as recurrent input to the model.
\end{itemize}
\begin{figure}[ht]
    \centering
    \begin{tikzpicture}
    \node (center) at (0,0) {};

    \node (hvector) at ($(center)+(0,0)$) {\large{$\mathbf{h} = $}\normalsize};
    
    \node (inneuron1) [draw, circle, minimum size=20, inner sep=0.0, red, fill=red!50,] at ($(hvector)+(1,0)$) {\textcolor{black}{$\mathbf{u}_1$}};
    \node (inneuron2) [draw, circle, minimum size=20, inner sep=0.0, red, fill=red!50,] at ($(inneuron1.east)+(0.5,0.)$) {\textcolor{black}{$\mathbf{u}_2$}};
    \node (inneuron3) [draw, circle, minimum size=20, inner sep=0.0, red, fill=red!50,] at ($(inneuron2.east)+(0.5,0.)$) {\textcolor{black}{$\mathbf{u}_3$}};
    \node (inneuron4) [draw, circle, minimum size=20, inner sep=0.0, red, fill=red!50,] at ($(inneuron3.east)+(0.5,0.)$) {\textcolor{black}{$\mathbf{u}_4$}};
    \node (fbneuron1) [draw, circle, minimum size=20, inner sep=0.0, orange, fill=orange!50,] at ($(inneuron4.east)+(0.5,0.)$) {\textcolor{black}{$\mathbf{v}_1$}};
    \node (fbneuron2) [draw, circle, minimum size=20, inner sep=0.0, orange, fill=orange!50,] at ($(fbneuron1.east)+(0.5,0.)$) {\textcolor{black}{$\mathbf{v}_2$}};
    \draw[line width=0.5] ($(inneuron1)+(-0.35, 0.55)$) -- ($(inneuron1)+(-0.5, 0.55)$) -- ($(inneuron1)+(-0.5, -0.55)$) -- ($(inneuron1)+(-0.35, -0.55)$);
    \draw[line width=0.5] ($(fbneuron2)+(0.35, 0.55)$) -- ($(fbneuron2)+(0.5, 0.55)$) -- ($(fbneuron2)+(0.5, -0.55)$) -- ($(fbneuron2)+(0.35, -0.55)$);
\end{tikzpicture}
    \caption{Structure of the internal state vector \( \mathbf{h}(t) \), with segmentation neurons \( \mathbf{u}_1, \dots, \mathbf{u}_4 \), and feedback neurons \( \mathbf{v}_1, \mathbf{v}_2 \). In this example $l=4$ and $k=2$.}
    \label{fig:hidden_state_vector}
\end{figure}
The full hidden state is \( \mathbf{h}(t) = [\mathbf{u}(t), \mathbf{v}(t)] \), where \( d = l + k \). To support predictive coding, we set $k$ to the number of classes, ensuring that the number of feedback neurons matches the number of semantic classes. This allows the model to compute class specific prediction errors and propagate them backward through the network.

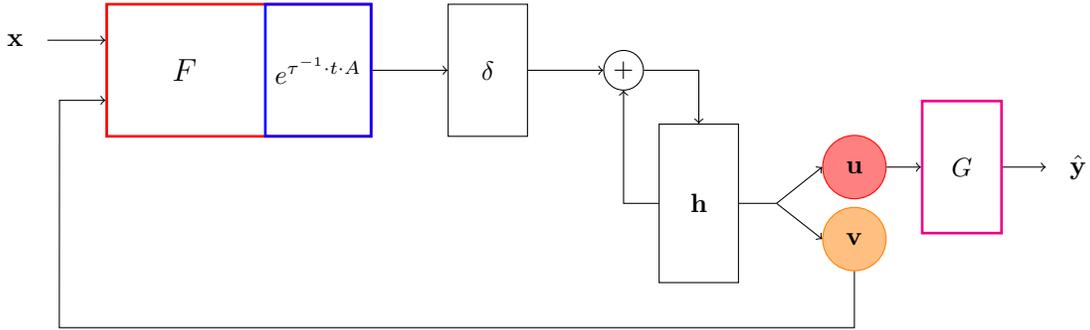
\begin{figure}[ht]
    \centering
    \begin{tikzpicture}

    \node (center) at (0,0) {};

    \node (F) [draw, rectangle, line width=1., minimum width=100, minimum height=50, inner sep=0., red] at ($(center)+(0,0)$) {};
    
    \node (expdecay) [anchor=east, draw, rectangle, line width=1., minimum width=40, minimum height=50, inner sep=0., blue] at ($(F.east)+(0,0)$) {\textcolor{black}{$e^{\tau^{-1}\cdot t \cdot A}$}};

    \node (delta) [draw, rectangle, minimum width=30, minimum height=50, anchor=west] at ($(expdecay.east)+(1,0)$) {$\delta$};
    \node (Flegend) at ($(F.west)!0.5!(expdecay.west)$) {\large{\textcolor{black}{$F$}}\normalsize};

    \node (sum) [anchor=west, draw, circle, minimum size=15, inner sep=0.] at ($(delta.east)+(1,0)$) {$+$};
    
    \node (h) [draw, rectangle, minimum width=30, minimum height=60] at ($(sum.south)+(1,-1.5)$) {$\mathbf{h}$};
    \node (hsplit) [anchor=west, draw, rectangle, minimum width=30, minimum height=60, white] at ($(h.east)+(1,0)$) {};
    \node (hy) [anchor=north, draw, circle, minimum size=24, inner sep=0., red, fill=red!50] at ($(hsplit.north)+(0,-0.15)$) {\textcolor{black}{$\mathbf{u}$}};
    \node (hfb) [anchor=north, draw, circle, minimum size=24, inner sep=0., orange, fill=orange!50] at ($(hy.south)+(0,-0.1)$) {\textcolor{black}{$\mathbf{v}$}};

    \node (G) [draw, rectangle, line width=1., minimum width=30, minimum height=50, inner sep=0., magenta,] at ($(hy.east)+(1,0)$) {\textcolor{black}{$G$}};

    \node (y) [draw, circle, minimum size=24, inner sep=0., white] at ($(G.east)+(1,0)$) {\textcolor{black}{$\hat{\mathbf{y}}$}};

    \node (x) [draw, circle, minimum size=24, inner sep=0., white] at ($(F.west)+(-1.2,0.4)$) {\textcolor{black}{$\mathbf{x}$}};

    \draw[->] ($(hfb.south)+(0,0)$) -- ($(hfb.south)+(0,-0.75)$) -- ($(hfb.south)+(-10.57,-0.75)$) -- ($(F.west)+(-0.61,-0.4)$) -- ($(F.west)+(0,-0.4)$);
    \draw[->] ($(x.east)+(0,0)$) -- ($(F.west)+(0,0.4)$);
    \draw[->] ($(expdecay.east)+(0,0)$) -- ($(delta.west)+(0,0)$);
    \draw[->] ($(delta.east)+(0,0)$) -- ($(sum.west)+(0,0)$);
    
    \draw[->] ($(h.west)+(0,0)$) -- ($(sum.south)+(0,-1.5)$) -- ($(sum.south)+(0,0)$);

    \draw[->] ($(sum.east)+(0,0)$) -- ($(sum)+(1,0)$) -- ($(h.north)+(0,0)$);

    \draw[] ($(h.east)+(0,0)$) -- ($(h)!0.5!(hsplit)$);
    \draw[->] ($(h)!0.5!(hsplit)$) -- ($(hy.west)+(0,0)$);
    \draw[->] ($(h)!0.5!(hsplit)$) -- ($(hfb.west)+(0,0)$);
    
    \draw[->] ($(hy.east)+(0,0)$) -- ($(G.west)+(0,0)$);
    \draw[->] ($(G.east)+(0,0)$) -- ($(y.west)+(0,0)$);
\end{tikzpicture}
    \caption{Overview of the predictive coding feedback loop. The internal state $\mathbf{h}(t)$ is split into segmentation neurons $\mathbf{u}(t)$ (red) and feedback neurons $\mathbf{v}(t)$ (orange). Feedback is passed by concatenating $\mathbf{v}(t)$ with the input image $\mathbf{x}$, which is used to compute the update $\delta(t)$.}
    \label{fig:circuitry}
\end{figure}

\noindent\textbf{Feedback dynamics.} The feedback mechanism refines the internal state over time, starting from an initial blank state \( \mathbf{h}(0) = \mathbf{0} \). At each timestep, the state is updated via:
\begin{equation}\label{equation:update}
    \mathbf{h}(t) = \mathbf{h}(t-1) + \delta(t-1),
\end{equation}
where the prediction error \( \delta(t) \) is estimated using a neural network \( F \) applied to the input image and the feedback state:
\begin{equation}
    \delta(t) = \tau \frac{d \mathbf{h}}{dt} = F([\mathbf{x}, \mathbf{v}(t)]) \cdot e^{\tau^{-1}\cdot -t \cdot A}.
    \label{equation:exp_decay}
\end{equation}
The exponential decay term ensures that feedback updates diminish over time, leading the system to converge. In practice, $A$ is decomposed into eigenvectors and eigenvalues, as \( A = Q \cdot \Sigma \cdot Q^{-1} \). This makes the matrix exponential $e^{A} = Q \cdot e^{\Sigma} \cdot Q^{-1}$. The eigenvalues are set to \( \Sigma = -I \), guaranteeing negative eigenvalues for all modes. $Q$ and $Q^{-1}$ are learned via backpropagation.

\noindent\textbf{Stability mechanisms.} To prevent instability in the feedback loop, we apply two constraints:
\begin{itemize}
    \item \textbf{Softmax projection:} To keep feedback values bounded and interpretable, we normalize the feedback neurons using:
    \begin{equation}\label{equation:softmax}
        \mathbf{v}(t) = \text{softmax}(\mathbf{h}_{l:l+k}(t)),
    \end{equation}
    ensuring \( \mathbf{v}(t) \in \Delta^k \), the $k$-dimensional simplex.
    \item \textbf{Exponential decay:} The error signal is decayed over time using the matrix exponential, encouraging the model to converge to a stable internal representation, as hypothesized in predictive coding theory.
\end{itemize}

\noindent\textbf{Prediction and optimization.} At any timestep, the segmentation prediction is given by:
\begin{equation}
    \hat{\mathbf{y}}(t) = G(\mathbf{u}(t)).
\end{equation}
Given a ground truth segmentation mask \( \mathbf{y} \), the model is trained by minimizing the cross entropy loss
\begin{equation}
    \mathcal{L}(\mathbf{y}, \hat{\mathbf{y}}(t)) = -\mathbf{y} \log(\hat{\mathbf{y}}(t)).
\end{equation}
The state of the model at step $T$ is the sum of the initial state with the errors of each iteration. By applying BPTT on this state, we accumulate gradients across all steps:
\begin{equation}
    \nabla_F \mathcal{L} = \sum_t \nabla_F \mathcal{L}(\mathbf{y}, F([\mathbf{x}, \mathbf{v}(t)]).
\end{equation}
An ablation study confirms that the performance gains stem from the feedback mechanism itself, not just the accumulation of gradients over time.

\noindent\textbf{Architecture: Feedback U-Net.} The function \( F \) is implemented using a modified U-Net \cite{ronneberger2015u}. The segmentation head \( G \) is applied only to the \( \mathbf{u}(t) \) subset of the internal state. We expand the input and output channels of the network to support the feedback mechanism and use the last hidden layer before the segmentation head to emit the full state \( \mathbf{h}(t) \).

\section{Experimental Setting}
\label{section:setting}

We designed two experiments to evaluate the robustness and generalization of the feedback model:
\begin{enumerate}
    \item \textbf{Noise experiment:} assess how performance degrades with increasing levels of Gaussian noise added to the input;
    \item \textbf{Training set size experiment:} evaluate how the model generalizes when trained with very limited labeled data.
\end{enumerate}
Both experiments were performed using an artificial dataset. We generated a synthetic segmentation dataset, $\mathcal{D} = \bigcup_i^D \{ \mathbf{x}_i, \mathbf{y}_i \}$, composed of binary images, $\mathbf{x}_i \in \{ 0,1 \}^{H\times W}$ containing a single irregular polygon per instance. Each polygon was created by randomly sampling vertices within a bounded 2D region, ensuring its center remains within the image dimensions. Each image has two semantic classes: background and polygon.

\noindent\textbf{Noise experiment.} In this setting, we fixed the training set size to \( D = 200 \), and added Gaussian noise, $\mathcal{N}(0,\sigma)$, with standard deviation \( \sigma \in \{0, 1, 2, \dots, 10\} \) to both training and test sets. The goal is to test how feedback and feedforward models behave under increasing input corruption.

\noindent\textbf{Train set size experiment.} Here, we fixed the noise level to \( \sigma = 0 \) (i.e., clean data) and varied the size of the training set as \( D \in \{1, 2, \dots, 10\} \). The test set size was fixed to 20 examples. This experiment assesses the ability of the model to generalize from few labeled examples.

\noindent\textbf{Hyperparameters.} Both feedback and feedforward models were trained using the Adam optimizer with a learning rate of \( 0.01 \), for 10 epochs, and a batch size of 1. The loss function used was cross entropy. The feedback model used a time constant of \( \tau = 1 \), as defined in equation~\ref{equation:exp_decay}.

\noindent\textbf{Evaluation metric.} We used the f1-score to evaluate segmentation performance, focusing on the polygon class (the minority class). The f1-score is computed as:
\begin{equation}
\text{f1-score} = 2 \times \frac{p \cdot r}{p + r + \epsilon},
\end{equation}
where \( p \) is precision and \( r \) is recall.

\section{Results}\label{section:results}

Table \ref{tab:ablations} shows how the error computation (Equations \ref{equation:exp_decay} and \ref{equation:update}) and the softmax operation applied to the feedback neurons (Equation \ref{equation:softmax}) affect both the feedback and feedforward models. Across all variants, the feedback model consistently outperforms its feedforward counterpart. These operations, particularly softmax and exponential decay, are not essential for performance per se, but rather for ensuring that the feedback system converges and remains stable over time. The softmax operation is not applicable to the feedforward baseline due to the lack of feedback. To ensure fairness, error computation in the feedforward model was approximated by setting \( t = 5 \), matching the final feedback timestep.

\begin{table}[ht]
    \centering
    \begin{tabular}{p{2cm} | p{2cm} |  p{2.cm} | p{2.cm}}
        \hline
        \hline
        \hfil \multirow{2}{*}{error $\delta(t)$} & \hfil \multirow{2}{*}{softmax $\mathbf{h}(t)$} & \multicolumn{2}{|c}{f1-score}  \\\cline{3-4}
        & & \hfil FF & \hfil FB \\
        \hline
        \hfil \multirow{2}{*}{\cmark} & \hfil \cmark  & \hfil \multirow{2}{*}{$0.32 \pm 0.03$} & \hfil $0.84 \pm 0.12$\\
        \hfil & \hfil \xmark  &  & \hfil $0.78 \pm 0.07$\\
        \hfil \multirow{2}{*}{\xmark} & \hfil \cmark  &  \hfil \multirow{2}{*}{$0.34 \pm 0.01$} & \hfil $0.84 \pm 0.04$\\
        \hfil & \hfil \xmark  &  & \hfil $0.84 \pm 0.06$\\
        \hline
        \hline
    \end{tabular}
    \caption{Ablation study on error computation and softmax feedback. Results are averaged across all noise levels. The feedforward model does not incorporate feedback, so softmax has no effect. \textit{Note:} To apply exponential decay in the feedforward model, we simulate a single-step decay using a static multiplicative factor \( e^{-\frac{t}{\tau} \cdot A} \), where \( t = 5 \) and $\tau=1$. This layer is not recurrent, but simply attenuates the output.}
    \label{tab:ablations}
\end{table}

\begin{figure}[ht]
    \centering
    \begin{tikzpicture}
        \node (noise) at (0,0) {\includegraphics[clip, trim={22.75cm, 0cm, 4cm, 0cm}, width=0.45\textwidth]{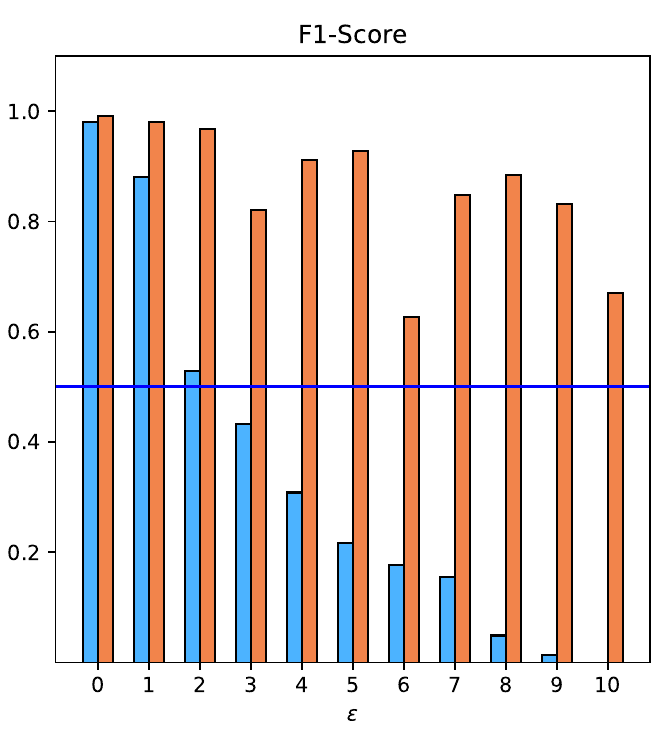}};
        \node (examples) [anchor=west] at ($(noise.east)+(0.1,0)$) {\includegraphics[clip, trim={22.75cm, 0cm, 4cm, 0cm}, width=0.45\textwidth]{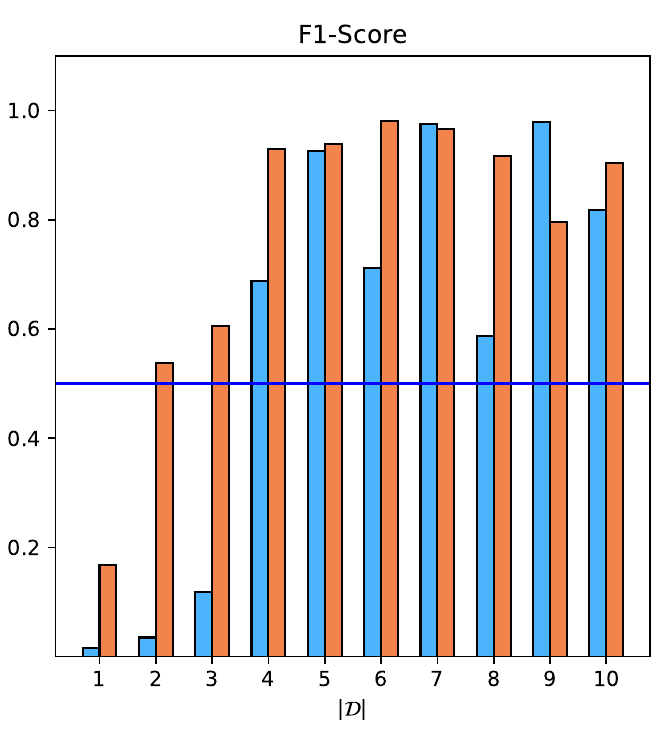}};
        \node (legend) at ($(noise.north)!0.5!(examples.north)+(0,0.1)$) {\includegraphics[clip, trim={34cm, 6cm, 0cm, 5cm}, width=0.2\textwidth]{results/figures/examples.pdf}};
        \node [draw, rectangle, minimum width=100, minimum height=15, white, fill=white, anchor=north] at ($(noise.north)+(0,-0.1)$) {};
        \node [draw, rectangle, minimum width=100, minimum height=15, white, fill=white, anchor=north] at ($(examples.north)+(0,-0.1)$) {};
        \node [draw, rectangle, minimum width=15, minimum height=180, white, fill=white, anchor=west] at ($(examples.west)+(0,0.)$) {};
        \node [anchor=center, rotate=90] at ($(noise.west)+(-0.1,0.)$) {f1 score};

        \node (legendnoise) [draw, rectangle, minimum height=10, minimum width=10, white, fill=white, anchor=south] at ($(noise.south)+(0.18,0.23)$) {\textcolor{black}{$\noise{}$}};
        \node (legendexamples) [draw, rectangle, minimum height=10, minimum width=10, white, fill=white, anchor=south, inner sep=0] at ($(examples.south)+(0.18,0.30)$) {\textcolor{black}{$D$}};
    \end{tikzpicture}
    \caption{Left: model performance under increasing noise levels. Right: performance under low data training. Feedback consistently outperforms feedforward baselines in both conditions. Random performance is denoted with the blue horizontal line.}
    \label{fig:feedforward_vs_feedback}
\end{figure}

\noindent\textbf{Impact of noise and training size.} Figure \ref{fig:feedforward_vs_feedback} shows performance trends across the full range of noise levels and training set sizes. As noise increases, the feedforward model quickly deteriorates to near-random performance, while the feedback model remains well above baseline. In the low-data regime, the feedback model requires only two training examples to outperform random, while the feedforward model requires at least four.

\begin{figure}
    \centering
    \includegraphics[width=\textwidth]{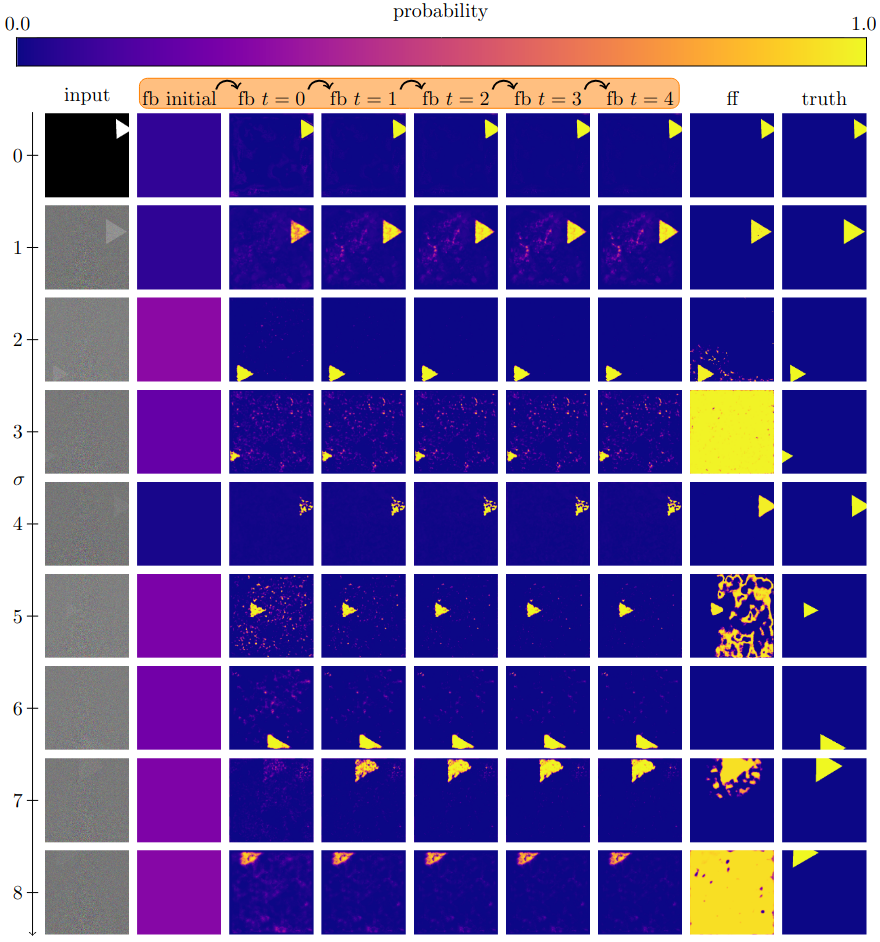}
    \caption{Noise impact on the prediction of the feedback versus the feedforward model. Feedback shows robustness across timesteps, especially at higher noise levels.}
    \label{fig:noise_feedback_vs_feedforward}
\end{figure}

\noindent\textbf{Robustness to noise.} In Figure~\ref{fig:noise_feedback_vs_feedforward}, we compare how the two models respond to increasing levels of Gaussian noise. At \( \noise{} = 0 \), both models perform similarly. However, as noise increases, the feedforward model rapidly degrades, while the feedback model retains meaningful structure. This highlights the feedback model’s robustness under corrupt input, which it achieves through iterative error correction over time.

\begin{figure}[ht]
    \centering
    \includegraphics[width=1.\linewidth]{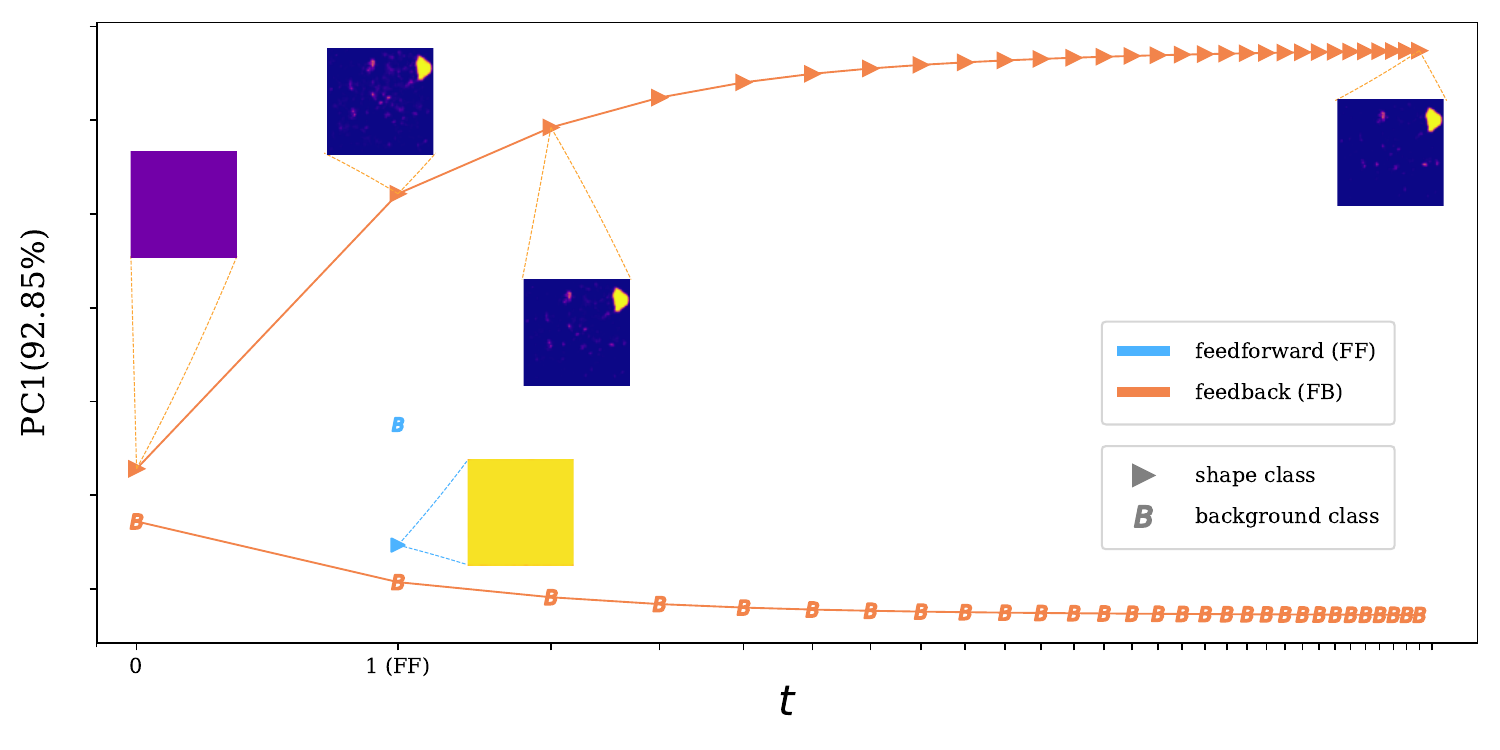}
    \caption{Temporal evolution of the strongest principal component (explains 90\% of variance) for the feedback model (across timesteps) and the feedforward model (single prediction). Feedback logits stabilize quickly, indicating convergence of internal state.}
    \label{fig:pca_feedback}
\end{figure}

\noindent\textbf{PCA analysis of internal dynamics.} To assess the stability of internal representations, we analyze the trajectory of the state vector over time. Specifically, we extract the logit subset \( \tilde{\mathbf{h}}(t) \) for all timesteps and test instances (at \( \noise{} = 6 \)), stack them into a matrix, and project onto the strongest principal component (PC). This projection reveals the dominant mode of variation in the high-dimensional logits over time. Importantly, this PCA does not evaluate accuracy, but captures whether the internal representation converges to a fixed point. A stable model should show little change across timesteps in this projection. This approach is conceptually related to linear classifier probes \cite{alain2016understanding}, which aim to reveal structural properties of intermediate network layers. Figure \ref{fig:pca_feedback} shows that the feedback model quickly settles into a stable representation, while the static feedforward model remains fixed. Despite initial variation in logits, the feedback system stabilizes after just a few steps, consistent with predictive coding dynamics. The feedforward model, on the other hand, misclassifies all pixels as shape class, showing no discrimination between object and background.

\begin{figure}[ht]
    \centering
    \includegraphics[width=1.\linewidth]{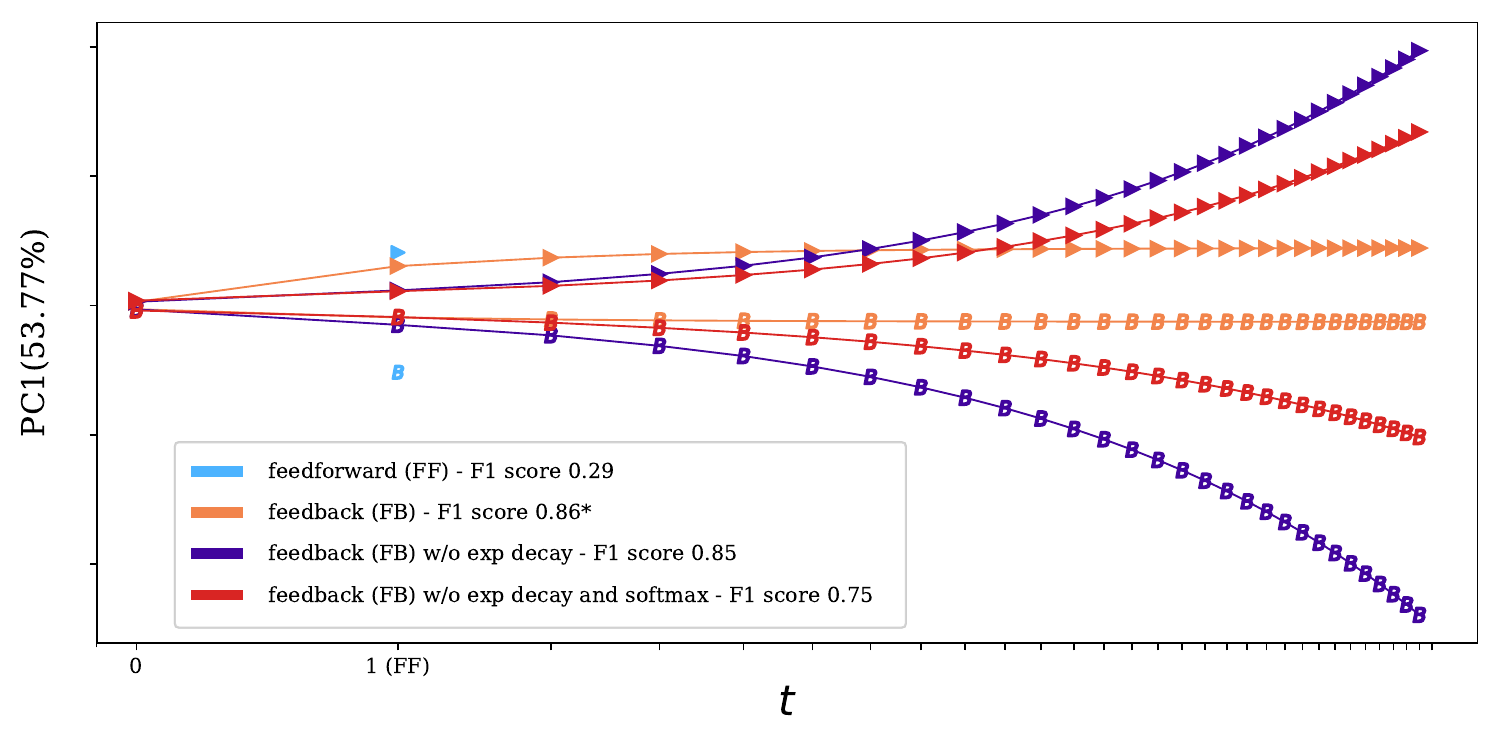}
    \caption{PCA trajectory without exponential decay (top) and without both decay and softmax (bottom). While f1-score remains high, the internal states diverge, indicating instability.} 
    \label{fig:pca_various_feedback}
\end{figure}

\noindent\textbf{Effect of removing stability mechanisms.} In Figure~\ref{fig:pca_various_feedback}, we repeat the PCA analysis on variants of the feedback model where stability mechanisms are removed. When exponential decay is omitted, the internal state diverges over time, even if the final prediction remains accurate. Removing both exponential decay and softmax further destabilizes the trajectory. These results demonstrate that softmax projection and exponential decay are essential not for raw accuracy, but for enforcing convergence in the internal state, a key property of predictive coding systems.

\section{Discussion}\label{section:discussion}

\noindent \textbf{Learning a trajectory without supervision.}  
Our feedback model bears similarity to diffusion models \cite{ho2020denoising}, in that it constructs a trajectory from an initial (blank) state to a goal state (ground truth). However, unlike diffusion, where the trajectory is predefined and known, our model learns this path implicitly through internal iterations. This trajectory emerges from a sequence of internal corrections that reduce prediction error over time. In this sense, the feedback mechanism enables a kind of thought process or temporal reasoning, allowing the model to refine its internal state step by step, something unavailable to feedforward models, which are limited to a single transformation from input to output. As shown in our results, this iterative refinement gives feedback a distinct advantage, particularly in challenging scenarios.

\noindent \textbf{Stability through structure.}  
Although we have not formally proved the stability of the feedback system, our empirical analysis suggests that it is asymptotically stable. As visualized in Figure \ref{fig:pca_feedback}, the internal representations converge rapidly along the principal component direction, with minimal variance after a few steps. Importantly, ablations in Figure \ref{fig:pca_various_feedback} demonstrate that this stability is achieved only when both exponential decay and softmax projection are applied. Without them, the system diverges despite achieving competitive f1-scores. This shows that stability is not necessary for accuracy on a single instance, but it is crucial for robustness and reliable long term behavior. 

\noindent \textbf{Robustness to noise.}  
One of the key benefits of feedback is its robustness to input corruption. Our results show that, under increasing levels of Gaussian noise, the feedback model maintains a high f1-score (Figure \ref{fig:feedforward_vs_feedback}) and produces qualitatively accurate segmentations (Figure \ref{fig:noise_feedback_vs_feedforward}). The feedforward model, in contrast, degrades rapidly. These results echo human like robustness in object recognition under noise \cite{dodge2017study}, previously documented in behavioral studies \cite{dodge2017study,rusak2020increasing}. The feedback loop allows the model to adaptively refine its internal state, filtering out noise and aligning predictions more closely with the underlying signal. This aligns with findings in neuroscience, where top down signals from high level cortical areas (e.g., prefrontal cortex) modulate early visual processing \cite{hupe1998cortical}, reinforcing the biological plausibility of our approach.

\noindent \textbf{Few-shot generalization.}  
Feedback also enhances generalization in low data regimes. When trained on only a few labeled examples, the feedback model achieves non trivial performance, outperforming feedforward baselines even with as few as two examples (Figure \ref{fig:feedforward_vs_feedback}). This is particularly promising for domains like medical imaging, where labeled data is scarce. Our findings suggest that feedback not only improves robustness but also promotes learning from minimal supervision, possibly due to its ability to internally reprocess and refine partial knowledge over multiple steps.

\noindent \textbf{Limitations and outlook.}  
Unlike many prior works that introduce recurrence only at the final layer, our model defines an entire neural architecture (a U-Net) as a recurrent system. This design introduces additional complexity: longer inference times, and sensitivity to the choice of internal dynamics. The PCA analysis shows that, on small-scale datasets, the internal state of the feedback model appears to converge toward accurate decisions, even without explicit constraints on the feedback dynamics. However, as Figure \ref{fig:pca_various_feedback} illustrates, the continuous evolution of the internal state becomes problematic at scale. Without the stabilizing effects of exponential decay and softmax normalization, the feedback loop lacks a mechanism to dampen updates, causing the state to continually fluctuate instead of settling. This prevents the system from reaching equilibrium, ultimately leading to divergence, manifested as overflow or underflow during training, when applied to more complex datasets such as ImageNet. These symptoms highlight the critical role of stability mechanisms in enabling reliable long term feedback dynamics.

\section{Conclusion}

We presented a novel feedback mechanism for neural networks that routes information from the output layer back to the input, enabling internal error correction over time. This architecture, grounded in predictive coding theory, departs from prior works by integrating feedback not just within layers but across the full depth of the network, from output to input, without requiring a predefined trajectory. Through controlled experiments on a synthetic segmentation task, we demonstrated that this mechanism improves robustness to input noise and generalization under limited supervision. We also introduced stability inducing operations, softmax projection and exponential decay, that ensure convergence of the internal state. To our knowledge, this is the first study to explicitly implement full network feedback in this way. Our findings suggest that even simple architectural changes, when aligned with biological principles, can significantly enhance neural network performance under challenging conditions. We hope this work encourages further exploration of feedback as a core design principle for more robust and adaptive artificial systems.

\section*{Acknowledgments}

This work was supported by the Center for Responsible AI project with reference C628696807-00454142, financed by the Recovery and Resilience Plan, and INESC-ID pluriannual UIDB/50021/-2020.

\bibliographystyle{unsrtnat}
\bibliography{bibliography}

\end{document}